\documentclass[letterpaper, 10pt, conference]{ieeeconf}  %

\IEEEoverridecommandlockouts                              %

\usepackage[T1, OT1]{fontenc}
\DeclareTextSymbolDefault{\dh}{T1}
\DeclareTextSymbolDefault{\DH}{T1}
\usepackage{times}

\usepackage{graphics,epsfig,graphicx,subcaption}
\usepackage{amsmath,amssymb,pifont,xcolor}
\usepackage{multicol,multirow,booktabs,arydshln}
\usepackage{hyperref,cite}

\newcommand{\cmark}{\ding{51}}
\newcommand{\xmark}{\ding{55}}

\newcommand{\map}[1]{\makebox{mAP50-95}#1}
\newcommand{\eg}{{\textit{e.g.}}}

\DeclareMathOperator{\argmax}{argmax}
\DeclareMathOperator{\softmax}{softmax}

\title{\LARGE \bf
Cross-Architecture Auxiliary Feature Space Translation for Efficient Few-Shot Personalized Object Detection %

}

\author{Francesco Barbato$^{1,2,*}$, Umberto Michieli$^{1}$, Jijoong Moon$^{3}$, Pietro Zanuttigh$^{2,\dag}$, Mete Ozay$^{1}$%
\thanks{*Research completed during internship at Samsung R\&D Institute UK.}
\thanks{$^\dag$ This work was partially supported by the European Union under the Italian National Recovery and Resilience Plan (NRRP) of NextGenerationEU, partnership on ``Telecommunications of the Future'' (PE00000001 - program ``RESTART'').}%
\thanks{$^{1}$Samsung R\&D Institute UK (SRUK),
        Communications House, South St, Staines, Surrey, United Kingdom
        {\tt\small \{u.michieli,m.ozay\}@samsung.com}}%
\thanks{$^{2}$University of Padova, via Gradenigo 6/b, 35131, Padova, Italy.
        {\tt\small \{francesco.barbato,zanuttigh\}@dei.unipd.it}}%
\thanks{$^{3}$Samsung Research Korea,
        Seoul R\&D Campus, 56, Seongchon-gil, Seocho-gu, Seoul, Rep. of Korea
        {\tt\small jijoong.moon@samsung.com}}%
}

\begin{document}

\maketitle
\thispagestyle{empty}
\pagestyle{empty}

\begin{abstract}
Recent years have seen object detection robotic systems deployed in several personal devices (e.g., home robots and appliances). 
This has highlighted a challenge in their design, i.e., they cannot efficiently update their knowledge to distinguish between general classes and user-specific instances (e.g., a \textit{dog} vs. \textit{user's dog}). 
We refer to this challenging task as \textit{Instance-level Personalized Object Detection} (IPOD). 
The personalization task requires many samples for model tuning and optimization in a centralized server, raising privacy concerns. An alternative is provided by approaches based on recent large-scale Foundation Models, but their compute costs preclude on-device applications.

In our work we tackle both problems at the same time, designing a \textit{Few-Shot IPOD} strategy called AuXFT. We introduce a conditional coarse-to-fine few-shot learner to refine the coarse predictions made by an efficient object detector, showing that using an off-the-shelf model leads to poor personalization due to neural collapse. 
Therefore, we introduce a \textit{Translator} block that generates an auxiliary feature space where features generated by a self-supervised model (e.g., DINOv2) are distilled without impacting the performance of the detector.
We validate AuXFT on three publicly available datasets and one in-house benchmark designed for the IPOD task, achieving remarkable gains in all considered scenarios with excellent time-complexity trade-off: AuXFT reaches a performance of 80\% its upper bound at just 32\% of the inference time, 13\% of VRAM and 19\% of the model size.
\end{abstract}

\section{INTRODUCTION}
Object Detection (OD) systems have been a staple in robotics and on a variety of personal devices such as home appliances (\eg, robot vacuum cleaners) and personal assistants \cite{sharma2019internet} for many years. 
Indeed, several applications require precise localization and recognition of the surrounding objects (\eg, for navigation or object grasping).
While most of them need only
coarse-level classes (e.g., \textit{person}, \textit{door}, \textit{plant}, etc.), emerging user-centric scenarios demand fine-grained detection abilities to identify personal instances (\eg, \textit{user's friend}, \textit{kitchen door}, \textit{chamomile}).
For example, a user watching the television may ask an assistive robot to retrieve its remote, rather than the remote for the \makebox{hi-fi}; a robot smart vacuum cleaner to clean the floor in front of a specific piece of furniture; a lawn mower to avoid specific plants or flowers; a phone to find photos of the user's pets, and so on.
We envision that OD systems should allow instance-level personalized understanding (\textit{e.g.}, \textit{dog} vs. \textit{user's dog}) when deployed on resource-constrained robotic devices through a very limited set of user-provided reference samples.
We define this task as \textit{Few-Shot Instance-level Personalized Object Detection} (FS-IPOD). 

FS-IPOD presents various challenges, chief among which are privacy concerns %
related to personal data processing. 
Viable solutions, indeed, should update the OD system on the device itself,
while learning with a few samples provided by the user for training \cite{snell2017prototypical,9341140,9981461,10342477,paramonov2024swiss}.%
\begin{figure}[t]
    \centering
    \includegraphics[width=\linewidth]{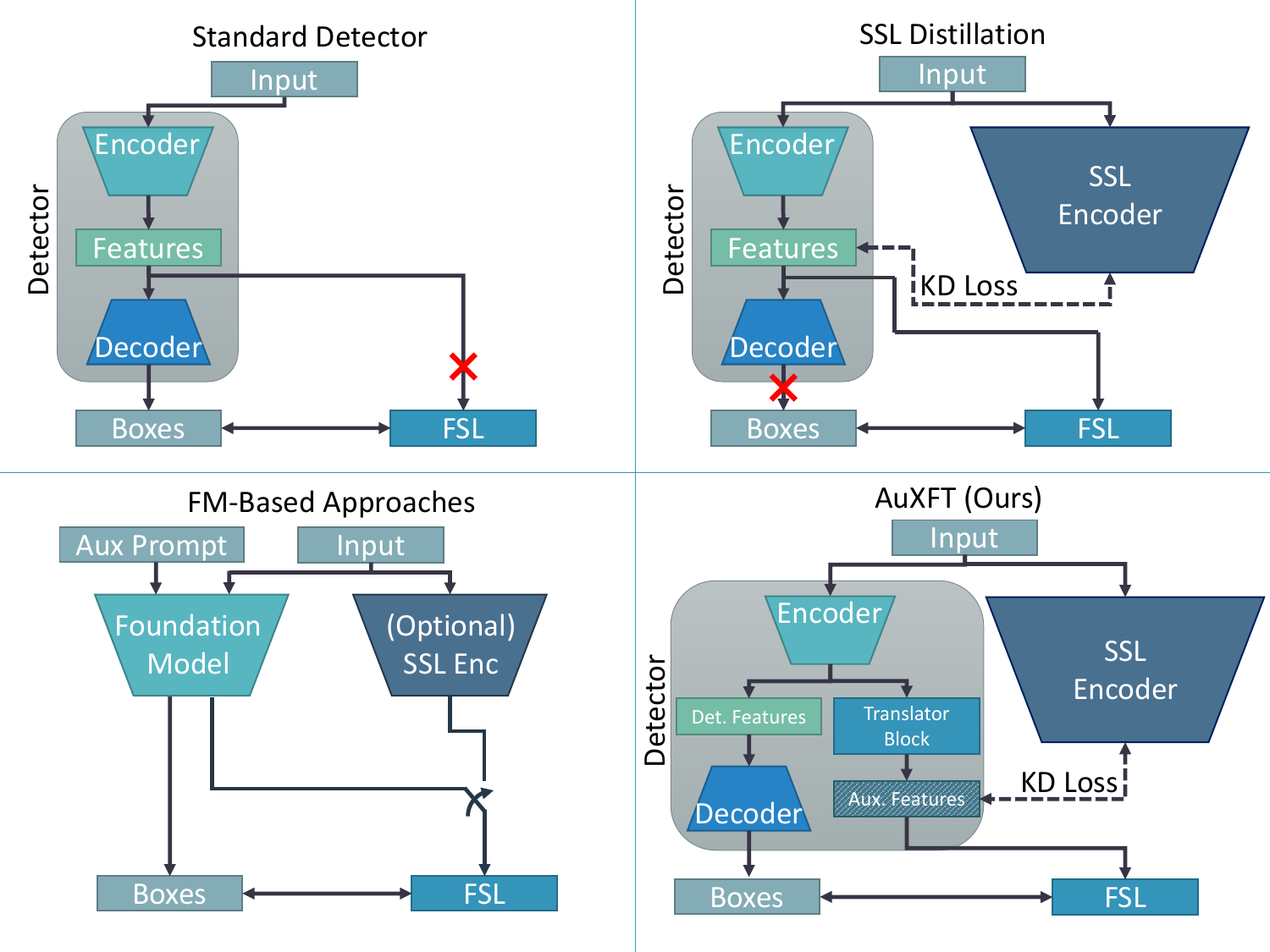}
    \caption{We explore few-shot instance-level personalized object detection in constrained robotic applications. 
    \textbf{Top-left:} standard pre-trained models suffer from \textit{neural collapse}, hence the FSL personalization fails. \textbf{Top-right:} na\"ive knowledge distillation degrades detection performance significantly. \textbf{Bottom left:} Foundation Model-driven approaches use compute-heavy vision-text architectures as guidance for feature pooling. \textbf{Bottom right (ours):} we create an auxiliary feature space where teacher knowledge is distilled for FSL personalization, without impacting detection performance. %
    \vspace{-1em}}
    \label{fig:graph_abs}
\end{figure}
Regrettably, deep learning based OD systems cannot efficiently update their knowledge once pre-training is complete, and fine-tuning them is typically very computationally expensive.
To cope with this, several post-training few-shot learning (FSL) strategies have been proposed \cite{snell2017prototypical,wang2019simpleshot,ziko2020laplacian}. These approaches, however, %
require the features processed by the architecture %
to be descriptive enough to distinguish between instances. 

As summarized in Fig.~\ref{fig:graph_abs}, existing approaches fall short in learning both discriminative (for %
OD%
) and descriptive (for %
personalization%
) features. %
Standard OD systems \cite{yolov8_ultralytics} suffer from the \textit{Neural Collapse} (NC) phenomenon \cite{papyan2020prevalence,tiomoko2024deep}. More precisely, %
the cross-entropy (CE) objective at pre-training stage causes the models' features to collapse around class centroids, losing intra-class variance and hence their usefulness for the personalization stage.
More descriptive features can be learned by models pre-trained via self-supervised learning (SSL) losses \cite{jaiswal2020survey} %
instead of CE, 
making these models especially suited as feature extractors for FSL modules.
Nonetheless, distilling knowledge from an SSL pre-trained model (\textit{e.g.}, DINO \cite{caron2021emerging,oquab2023dinov2}), into an OD system entails several compatibility challenges only recently addressed \cite{Liu_2022_ACCV,zhao2023cross,hao2024one} and can affect features for a downstream OD task.
Other works explore %
personalization of large semantic segmentation Foundation Models (\textit{e.g.}, SAM \cite{kirillov2023segment}); such approaches require multiple architectures and, therefore, are not suitable for resource-constrained applications~\cite{zhang2024personalize,10341474}.

To restore descriptiveness in the features, we propose a cross-architecture knowledge transfer approach via an auxiliary feature space called AuXFT.
It aims at improving the personalization ability of OD models while keeping the detection performance unaltered. To achieve this, we distill the knowledge of an SSL-pretrained model %
into an %
auxiliary feature space of a detector.
However, construction of the auxiliary feature space is highly nontrivial due to the intrinsic architectural differences between the models of interest: in our case, a convolutional network (e.g., YOLOV8n \cite{yolov8_ultralytics}), and a vision transformer (e.g., DINOv2 \cite{oquab2023dinov2}). 
The two architectures have different numbers of feature maps, channel depths, and resolution layers. To align the feature spaces of the models,  we propose a \textit{translator block} which exploits two \textit{differential} modules.
We evaluate the effectiveness of our approach for 1-shot and 5-shot tasks on 3 public datasets (PerSeg \cite{zhang2024personalize}, iCubWorld \cite{pasquale2015teaching}, and CORe50 \cite{lomonaco2017core50}), as well as on an in-house benchmark called Personalized Object Detection (POD), capturing personal objects in indoor scenes.
\footnote{Code and data are available at:\\\url{https://github.com/SamsungLabs/AuXFT}.}. 
AuXFT shows significant improvements in all scenarios. 

The major contributions of our work are: 
i) We design a pipeline to personalize object detectors to instance-level classes by creating an auxiliary feature space that is used for %
personalization%
, while keeping the detection inference unaltered; 
ii) we introduce a knowledge translator block to align the %
detector's auxiliary feature space to %
the one of an SSL feature extractor, allowing knowledge distillation losses to bring intra-class variability into the auxiliary feature space;
iii) we design a conditional coarse-to-fine FSL block that extracts representative embedding vectors from the auxiliary space via pooling and employs them for personalized instance-level retrieval with minimal computation. %

The rest of the manuscript is organized as follows: Sec.~\ref{sec:related} summarizes the related works; %
Sec.~\ref{sec:problem} outlines the problem statement; Sec.~\ref{sec:method} introduces each component of our approach; Sec.~\ref{sec:results} reports the experimental results, whose analysis is extended in Sec.~\ref{sec:ablation};
and finally, Sec.~\ref{sec:conclusions} concludes and discusses possible future venues of research.

\begin{figure*}[t]
    \centering
    \includegraphics[width=.9\textwidth]{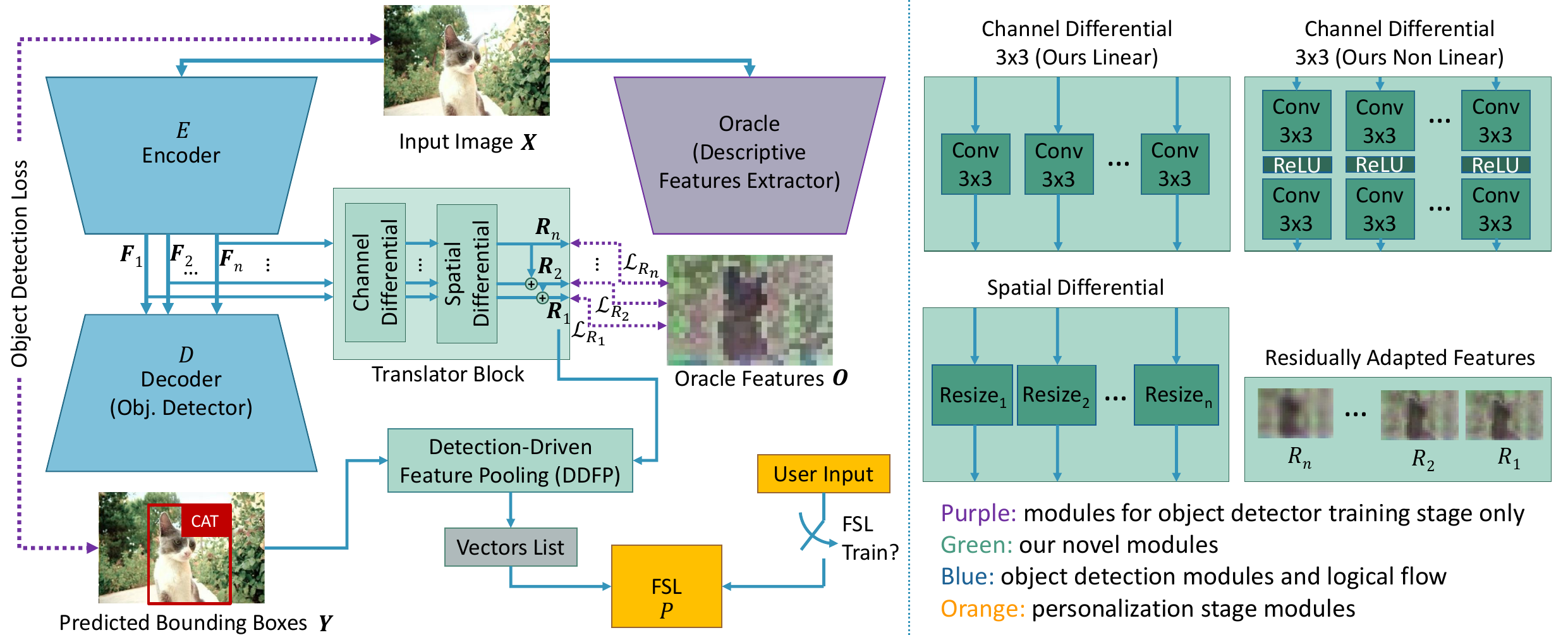}
    \caption{Overview of our AuXFT. The descriptive features produced by the oracle network are distilled in the auxiliary space through the Translator Block during training. At inference time, the oracle is discarded. For personalization, the predicted boxes and features pass through the DDFP, whose output is fed to the FSL. User input is only needed for FSL training. %
    \vspace{-1em}}
    \label{fig:arch}
\end{figure*}

\section{Related Works}\label{sec:related}
\textbf{Cross-Architecture Knowledge Distillation }
has emerged in recent years as a generalization of hint-based knowledge distillation approaches \cite{romero2014fitnets,zhang2020task,passalis2020heterogeneous,luo2019knowledge} (which assume that the architectures' topology matches). The objective is to distill transformer architectures \cite{vaswani2017attention,dosovitskiy2020image,devlin2019bert,liu2021swin,wang2021pyramid,barbato2023depthformer,rizzoli2022multimodal} into more efficient alternatives (\eg, CNNs), since their impressive performance comes at a significant compute cost.
The first investigation into the topic by Liu et al.\ \cite{Liu_2022_ACCV} focused on distilling the output space of two heterogeneous architectures. %
In \cite{zhao2023cross} the authors could improve the performance of a face recognition model %
introducing an ad-hoc distillation approach that used facial keypoints as hints and focused on the attention maps produced by the teacher.
The most recent work on the topic, which is also the closest to our approach, is \cite{hao2024one} where distillation on outputs and feature spaces of models is applied simultaneously, mimicking shortcut-based architectures. 
The core differences compared to our approach lie in the utilized distillation techniques and overall objective. \cite{hao2024one} focuses on achieving the highest accuracy on a task shared by teacher and student models, via KL-divergence-based losses to distill teacher logits into the student's output space and projected features. 
Our objective is different since we consider the reconstruction of the teacher's feature space as an auxiliary task, which we want to solve without impacting the performance of the main task. Therefore, we use losses more geared toward reconstruction and apply them to an auxiliary, translated space.

\textbf{Few Shot Learning }
task was first introduced by Miller et al.\ \cite{miller2000learning} and it has been extensively investigated in the past years. 
The seminal work by Snell et al.\ \cite{snell2017prototypical}  sparked a variety of approaches \cite{wang2019simpleshot,ziko2020laplacian}. These techniques are based on a prototypical classification approach where class-representative vectors (i.e., prototypes) are stored in memory and used to classify an unknown query vector. Recently, FSL has seen impressive development thanks to its potential in a variety of tasks beyond image classification, especially for dense scene understanding \cite{wang2019panet,10342477,9981461,9341140}.

\textbf{Personalized Scene Understanding }
task traces its roots in NLP \cite{mirkin2015motivating} where the conversation tone tends to change dramatically for different users, and architectures should adapt accordingly. The first to investigate its application to vision tasks was Zhang et al.\ \cite{Zhang_2021_ICCV}, which applied the concepts to semantic segmentation.
Since then, some few-shot techniques have been introduced, thanks to the development of Foundation Models (FM) such as CLIP \cite{radford2021learning} and SAM \cite{kirillov2023segment}, these architectures allow the use of textual prompts to guide the behavior of vision models. Some examples of this recent trend of research can be identified in PerSAM \cite{zhang2024personalize}, Matcher \cite{liu2024matcher}, and SegGPT \cite{wang2023seggpt}, where the core idea is to prompt an FM and use it to generate either segmentation masks or descriptive features, which can be used for the personalization.

\section{Problem Formulation}\label{sec:problem}
In this work, we focus on personalizing object detectors to instance-level classes for their significant real-world implications in various robot systems such as humanoid, service, and rescue robotics. %
Nonetheless, in principle, our approach can be applied in any scenario where the knowledge of a compute-heavy model needs to be distilled into a more efficient one, especially in the presence of a mismatch in the architectures or in the task performed by the networks.

In the FS-IPOD task, we assume that we have an OD model that recognizes coarse-level classes $\mathcal{C}_c$ (\textit{e.g.}, \textit{dog}, \textit{cat}, \textit{person}, etc.) and we want to expand the class set $\mathcal{C} = \mathcal{C}_c \cup \mathcal{C}_f$, to include classes more relevant to the user (\textit{e.g.}, \textit{dog1}, \textit{dog2}, \textit{cat1}, \textit{etc.}), without forgetting the original class set \cite{shenaj2022continual,barbato2023continual}.
This task is especially useful in personal devices (\textit{e.g.}, robotic assistants, smart vacuum cleaners, phones, \textit{etc.}) which come with very limited computational resources and where, due to the personal nature of the final class set, learning at server-side is undesirable. This means that the refinement operations must be performed in a compute-efficient manner and with a minimal amount of labeled samples since the end-user will need to provide them.

To address these issues, we resort to FSL, where a general model is trained on a chosen scenario with very few samples available for tuning (order of 1-5 samples per class). To mimic the limited number of samples, FSL evaluation is often performed in a cross-validation manner: in each \textit{episode} $\mathcal{T} = (\mathcal{T}_s, \mathcal{T}_q), \; \mathcal{T}_s \cap \mathcal{T}_q = \varnothing$, a \textit{support set} $\mathcal{T}_s$ of training samples is selected from the available samples, leaving the remaining as the validation \textit{query set} $\mathcal{T}_q$ where the performance is evaluated. This procedure is repeated a number of times, changing the support and query set each time, to obtain the average episodic accuracy, which is a more reliable estimate of the actual system performance.
When $|\mathcal{T}_s| = 1$ the scenario is referred to as 1-shot, when $|\mathcal{T}_s| = 5$ we refer to it as 5-shot.
A common approach to enable FSL is feature-level prototypical classification, which allows to split the processing burden in two. %
A frozen and compute-heavy feature extractor can be implemented in hardware, while the variable and efficient FSL classifier in software.

In this work, we propose AuXFT: an FS-IPOD method 
that can refine the original (\textit{i.e.}, coarse-level) detection class-set with a limited amount of user-labeled personal (\textit{i.e.}, fine-level) samples.
In object detection, the input space ${\mathcal{X} \subset \mathbb{R}^{H \times W \times 3}}$ can be identified as the set of  RGB images of size $H \times W$; while the output space is a list of coordinates ($x_0,y_0,x_1,y_1$), class-assignments ($c$), and classification confidence ($p$): ${\mathcal{Y} = \{(x_0,y_0,x_1,y_1,c,p)\}_{k = 1, \dots, K} \subset (\mathbb{R}^4 \times \mathcal{C}_c \times [0,1])^K}$; where $K$ is the number of predicted detections.
To enable FSL, we choose an encoder-decoder detector (in particular YOLOV8n \cite{yolov8_ultralytics}), \textit{i.e.} $M = D \circ E: \mathcal{X} \mapsto \mathcal{Y}$, where $\circ$ is the composition operator, so that we can access the intermediate features.
Given an input image $\mathbf{X} \in \mathcal{X}$, the encoder produces a set of features $\mathcal{F} \ni \{\mathbf{F}_1,\mathbf{F}_2, \dots \mathbf{F}_n\} = E(\mathbf{X})$ which we can feed to the FSL block $P: \mathcal{F} \mapsto \mathcal{C}$ to obtain the fine-grained classification. 

\section{Method}\label{sec:method}
Our system is based on an efficient object detector network, that is modified by our feature \textit{Translator Block}.
This block generates an auxiliary feature space where the knowledge of an oracle SSL feature extractor is distilled. The auxiliary features are used at the personalization stage to compute descriptive embeddings for the objects in the scene.
Note that the oracle model is only used at training time and discarded for inference.
A graphical representation of AuXFT is shown in Fig.~\ref{fig:arch}. %

\subsection{Translator Block}\label{subsec:translator}
We begin by describing our feature translator block, which we use to generate the auxiliary feature space.
The block consists of two major components: a \textit{Channel Differential} module ($D_C$) and a \textit{Spatial Differential} module ($D_S$) that will be detailed next. The two modules translate the original feature space into an intermediate space where all feature maps have the same resolution and channel depth. Such features are then used to produce the auxiliary features via residual chain by
\begin{equation}
    \mathbf{R}_i = \sum_{j=i}^{n} D_S(D_C(\mathbf{F}_j)) \quad\quad \forall i=1,2, \ldots,n.
\end{equation}
This allows the high-fidelity and low-resolution estimates generated by the channel-dense features to have their high-frequency components restored by their more shallow but higher-resolution counterparts.
We remark that only $\mathbf{R}_1$ is used at inference time by the \textit{Detection-Driven Feature Pooling} (Sec.~\ref{subsec:ddfp}) to generate the embedding vectors fed to the FSL block (Sec~\ref{subsub:fsl}). %

As common in reconstruction tasks \cite{fu2006efficient}, we employ the sum of $\ell_1$ and $\ell_2$ norms as the distillation loss by
\begin{equation}
    \mathcal{L}_{\mathbf{R}}%
    =  \sum_{i = 1}^{n}\sum_{p \in H'' \times W''}{\frac{\|\mathbf{R}_i[p] - \mathbf{O}[p]\|_1 + \|\mathbf{R}_i[p] - \mathbf{O}[p]\|_2}{H'' \times W''}} 
\end{equation}
where $\mathbf{O} = E_O(\mathbf{X})$ are the oracle features of an SSL %
extractor (e.g., DINOv2 \cite{oquab2023dinov2}) having resolution $H''\times W''$. %

\subsubsection{Channel Differential}\label{subsub:ch_diff}
In principle, the channel differential module can be any mapping $D_C$ from the feature space of the encoder to the feature space of the same spatial resolution, but with channel depth matching the one of the target oracle feature extractor. That is, ${D_C: \mathbb{R}^{H' \times W' \times {l}_E} \mapsto \mathbb{R}^{H' \times W' \times {l}_O}}$, where ${l}_E$ and $l_O$ are the number of channels of detector and oracle, respectively. As detailed in Sec. \ref{sec:results}, we tested a variety of different projections %
before %
choosing a mid-range option offering a good compromise between few-shot performance and computational cost, that is,
a $3\times3$ convolution. %

\subsubsection{Spatial Differential}\label{subsub:sp_diff}
After projecting the feature vectors in the teacher's channel space, we adapt the spatial dimensions via the spatial differential, to allow computation of the residual maps. 
In principle, this task can be performed by any mapping $D_S: \mathbb{R}^{H' \times W' \times {l}_O} \mapsto \mathbb{R}^{H'' \times W'' \times {l}_O}$.
For efficiency, we opted for a parameter-free approach using classical resampling algorithms.
Since the aspect ratio ($W\!\!:\!\!H$) of detector and oracle features is the same, we employed an adaptive strategy to select the interpolation algorithm depending on the resizing factor $\rho = \frac{H''}{H'} = \frac{W''}{W'}$ as %
\begin{equation}
    \text{Resampling alg.} = 
    \begin{cases}
        \text{i) area ,} & \text{if} \;\; \rho < 1-\delta\\
        \text{ii) bilinear,} & \text{if} \;\; 1-\delta \leq \rho < 1+\delta \\
        \text{iii) bicubic,} & \text{if} \;\; \rho \geq 1+\delta
    \end{cases}
\end{equation}
where $\delta$ is a parameter that we empirically set to $0.1$.
Such a choice finds root in classical signal processing \cite{mitra2001digital}: i) to downsample an image without introducing re-sampling artifacts the best strategy is to average the pixels from the high-resolution source; ii) if the target resolution is close to the source one, %
then there is no need for large interpolation kernels and a simple linear interpolation is enough, to reduce the complexity overhead; iii) to upsample an image of a larger factor, a more complex interpolation strategy (\textit{e.g.},  bicubic) is preferable to obtain more accurate results. %

\subsection{Detection-Driven Feature Pooling (DDFP)}\label{subsec:ddfp}
A fundamental component of our AuXFT is the re-labeling of the boxes detected by $M$. This is done using a prototype-based FSL (Sec.~\ref{subsub:fsl}), which requires embedding vectors as inputs. The task of converting the auxiliary feature $\mathbf{R}_1$ into such vectors (one for each bounding box, ${\mathcal{V} = \{\mathbf{v}_1, \mathbf{v}_2, \dots, \mathbf{v}_K\}}$) is handled by the DDFP block, that uses the predicted boxes to summarize the relevant regions of the map into a %
vector using spatial-aware average pooling:
\begin{align}
    \mathbf{v}_k = \frac{1}{x_1-x_0}&\frac{1}{y_1-y_0} \sum_{x = x_0}^{x_1-1} \sum_{y = y_0}^{y_1-1} \mathbf{R}_1[x,y], \\
    &\forall \;\; \mathbf{Y}_k = (x_0,y_0,x_1,y_1,c,p) \in M(\mathbf{X}). \nonumber
\end{align}
During FSL training, these vectors are accompanied by the original coarse class prediction $c \in \mathcal{C}_c$ and by a fine class assignment $c' \in \mathcal{C}_f$ provided by the user. At inference time, only the coarse class prediction is needed by the FSL module to predict a new fine-class assignment.

\begin{table*}[t]
\setlength{\tabcolsep}{3pt}
\centering
\caption{
Average \map{ }over 100 episodes ($\uparrow$). %
Our target is a good trade-off between performance (first four columns) and costs (last three columns). no pers.: metrics on coarse class set. Last two rows: AuXFT (3x3 lin.) score in relative terms compared to Yolo Only (last feat.) and Oracle (single fwd.), respectively.
$\dagger$ retrieval accuracy is used instead of \map. Note that the 5-shot results in POD do not have an std, because in that scenario all training samples are used for tuning. %
}
\label{tab:results:fine}
\resizebox{\textwidth}{!}{%
\begin{tabular}{cc|c|cc|cc|cc|c|c|c}
    \multirow{2}{*}{\rotatebox{90}{Arch.}} & \multirow{2}{*}{Setup} & PerSeg (13/39) & \multicolumn{2}{c|}{POD (15/45)} & \multicolumn{2}{c|}{iCubWorld$\dagger$ (6/31)} & \multicolumn{2}{c|}{CORe50$\dagger$ (10/50)} & Time  & VRAM & Size\\
    & & 1-shot & 1-shot & 5-shot & 1-shot & 5-shot & 1-shot & 5-shot & [ms/im] ($\downarrow$) & [MB] ($\downarrow$) & [MB] ($\downarrow$) \\
    \midrule
    \multirow{4}{*}{\rotatebox{90}{Standard}} & Yolo Only (last feat.) & 38.5\tiny{$\pm$2.7} & 20.9\tiny{$\pm$ 2.1} & 27.7 & 49.8\tiny{$\pm$2.4} & 65.2\tiny{$\pm$2.4} & 56.7\tiny{$\pm$6.5} & 65.6\tiny{$\pm$7.5} & 32.3\tiny{$\pm$5.2} & 221.4 & 12.2 \\ %
    & Yolo Only (concat.) & 41.2\tiny{$\pm$2.6} & 23.6\tiny{$\pm$2.4} & 30.4 & 51.2\tiny{$\pm$3.0} & 68.4\tiny{$\pm$2.4} & 57.8\tiny{$\pm$6.3} & 67.3\tiny{$\pm$7.7} & 50.5\tiny{$\pm$5.3} & 221.5 & 12.2 \\ %
    \cdashline{2-12}
    & Inverse \cite{zhang2020task} & 41.3\tiny{$\pm$3.0} & 21.1\tiny{$\pm$2.4} & 25.6 & 50.6\tiny{$\pm$2.9} & 63.9\tiny{$\pm$2.5} & 56.6\tiny{$\pm$6.0} & 65.4\tiny{$\pm$7.3} & 48.6\tiny{$\pm$5.3} & 221.5 & 12.2 \\ %
    & AvgStd \cite{10341474} & 41.0\tiny{$\pm$2.7} & 22.9\tiny{$\pm$2.2} & 27.8 & 50.6\tiny{$\pm$2.5} & 69.2\tiny{$\pm$2.1} & 58.0\tiny{$\pm$6.2} & 68.5\tiny{$\pm$7.1} & 53.5\tiny{$\pm$5.3} & 222.0 & 12.2 \\ %
    \midrule
    \multirow{4}{*}{\rotatebox{90}{Translated}} & AuXFT (1x1 lin.) & 49.6\tiny{$\pm$3.1} & 29.4\tiny{$\pm$2.5} & 35.4 & 55.1\tiny{$\pm$2.7} & 73.4\tiny{$\pm$2.1} & 58.4\tiny{$\pm$6.0} & 67.3\tiny{$\pm$7.6} & 32.7\tiny{$\pm$5.1} & 283.7 & 12.9 \\ %
    & AuXFT (3x3 lin.) & 48.8\tiny{$\pm$3.4} & 31.5\tiny{$\pm$2.7} & 38.8 & 55.0\tiny{$\pm$3.0} & 74.5\tiny{$\pm$2.5} & 58.8\tiny{$\pm$6.4} & 69.3\tiny{$\pm$6.8} & 32.6\tiny{$\pm$5.1} & 288.7 & 18.4 \\ %
    & AuXFT (5x5 lin.) & 49.5\tiny{$\pm$3.3} & 34.1\tiny{$\pm$2.5} & 39.1 & 55.0\tiny{$\pm$3.0} & 74.3\tiny{$\pm$2.3} & 59.3\tiny{$\pm$6.0} & 69.8\tiny{$\pm$7.1} & 34.6\tiny{$\pm$5.1} & 431.6 & 29.4 \\ %
    & AuXFT (3x3 non-lin.) & 50.4\tiny{$\pm$3.4} & 32.7\tiny{$\pm$2.3} & 40.5 & 52.3\tiny{$\pm$2.9} & 72.7\tiny{$\pm$2.5} & 59.8\tiny{$\pm$5.4} & 70.8\tiny{$\pm$5.9} & 36.1\tiny{$\pm$5.1} & 342.4 & 49.4 \\ %
    \midrule
    \multirow{2}{*}{\rotatebox{90}{Multi}} & Oracle (single fwd.) & 61.9\tiny{$\pm$3.0} & 40.4\tiny{$\pm$2.7} & 47.9 & 67.5\tiny{$\pm$2.6} & 84.8\tiny{$\pm$2.0} & 65.4\tiny{$\pm$7.6} & 75.7\tiny{$\pm$8.3} & 102.2\tiny{$\pm$4.7} & 2313.9 & 98.5 \\ %
    & Oracle (multiple fwd.) & 62.6\tiny{$\pm$3.1} & 46.0\tiny{$\pm$3.0} & 62.0 & 73.1\tiny{$\pm$2.6} & 87.9\tiny{$\pm$1.7} & 74.5\tiny{$\pm$6.3} & 84.5\tiny{$\pm$5.6} & 357.3\tiny{$\pm$12.8} & 338.3 & 98.5 \\ %
    \midrule[1pt]
    & Upper Limit (no pers.) & 69.9 & \multicolumn{2}{c|}{70.0} & \multicolumn{2}{c|}{99.8} & \multicolumn{2}{c|}{98.9} & - & - & -\\
    & AuXFT Rel.\ Baseline & 126.8\% & 150.7\% & 140.1\% & 110.4\% & 114.3\% & 103.7\% & 105.6\% & 100.9\% & 130.4\% & 150\% \\
    & AuXFT Rel.\ Oracle & 78.8\% & 78.0\% & 81.0\% & 81.5\% & 87.9\% & 89.9\% & 91.5\% & 31.9\% & 12.5\% & 18.7\% \\
\end{tabular}}
\vspace{-1em}
\end{table*}

\subsection{Few Shot Learner}\label{subsub:fsl}
Our FSL module is based on a prototypical network \cite{snell2017prototypical}, but is tailored for efficiency, given the computational constraints of robotic applications. In general, a prototypical network is trained by providing a series of labeled vectors, which are used to build internal class prototypes. %
During inference, the distance between a query vector and all internal prototypes is computed and used to classify the vector by choosing the class of the closest prototype.

This approach scales poorly with the number of classes. Therefore, we implement our FSL in a conditional fashion, that is, we limit the search space of the fine-level class on the basis of the coarse class predicted by the detector (rather than searching among all possible options). This allows our method to be about 3 times faster than the standard non-conditional FSL, without significant impact on the re-classification accuracy, as we observe in the ablation study.

Another difference compared to standard FSL modules is the introduction of an additional vector in each of the conditional sets, which we use as a fallback when a query does not match any fine class. It is the centroid of the set and will match the vectors that are not similar enough to any of the prototypes.
This addition requires the FSL to be robust and for this reason, we employed multiple distances and a majority voting approach for the classification.
Given a set of distance functions ${\mathcal{D} = \{d_1, d_2, \dots\}}$ , a query vector $\mathbf{q}$, its coarse class prediction $c$ and a set of prototypes $\mathcal{P}[c] = \{\mathbf{p}_1, \mathbf{p}_2, \dots\}$, we can compute a new class assignment as follows:
\begin{align}
    \mathbf{d} &= \frac{1}{|\mathcal{D}|}\sum_{d \in \mathcal{D}} \softmax \left\{ \frac{1}{d(\mathbf{q}, \mathbf{p}_i)}; \forall i = 1,\dots,|\mathcal{P}[c]|\right\}, \\
        c' &= \argmax\limits_{i = 1,\dots,|\mathcal{P}[c]|} \mathbf{d}[i].
\end{align}
Effectively, we compute an assignment probability distribution from the inverse of the distance between the query vector and the prototypes for each function (for the results we used cosine, $\ell_1$ and $\ell_2$ distances, see Sec.~\ref{sec:ablation}). We then average over the distances and select as the fine class the index corresponding to the maximum.

\subsection{Datasets}\label{subsub:datasets}
We pre-train all detectors on a subset of the OpenImages dataset \cite{OpenImages} with 31 coarse classes, covering the ones present in the personalization datasets, personalization classes absent from OpenImages were discarded. 

\textbf{PerSeg} \cite{zhang2024personalize} dataset contains 212 images in 39 instance-level classes (13 coarse). To the best of our knowledge, it is the only dataset proposed for few-shot personalized dense tasks.
PerSeg was originally proposed for semantic segmentation, so we converted the annotations for OD. %
Some instances have a limited number of samples (as low as 4), precluding the 5-shot scenario and the coarse-to-fine %
mapping is unbalanced, that is, some classes have only one personal example, while others have up to eight.

\textbf{POD} (Personalized Object Detection) dataset is an in-house benchmark of everyday objects in indoor scenes (\textit{e.g.}, glasses, shoes, bottles, etc.) developed to solve the limitations of PerSeg. 
We capture a total of 5 training images for each of the 45 instances considered (3 for each of the 15 coarse classes) for a total of 225 images. In the validation set, we capture images both in light and dark conditions. %
We acquired 3 validation images per instance: alone in the scene, mixed with the other same-class instances, and mixed with instances from different classes (for a total of 150 images).

\textbf{iCubWorld} \cite{pasquale2015teaching} dataset focuses on robotic views %
and has been acquired using video sequences. We select 15 samples per sequence (31 fine classes, 6 coarse) in the FSL split. %
The bounding boxes have noticeable padding around the objects %
making them incompatible with the %
pre-training. %
Therefore,
we fine-tuned the decoder before training the FSL and used the retrieval accuracy instead of OD metrics.

\textbf{CORe50} \cite{lomonaco2017core50} dataset %
provides a variety of object instances captured in variable-background scenes (10 classes, 5 instances per class). %
We selected a frame with a valid bounding box every thirty from the source videos. 
The boxes were generated algorithmically %
and are squared with significant padding around the object. Therefore, we fine-tune the detector and use the accuracy in %
this dataset as well. To align with the other domains, we set up the episodes on this dataset to have a single scene inside them. %

\subsection{Implementation Details}\label{subsub:impl}
We train using images (394,619 in total) with resolution 672x672px, Adam \cite{kingma2017adam} optimizer for 50 epochs adopting a cosine annealing with linear warmup (1000 iterations) scheduler for both learning rate (max $1\times10^{-3}$) and weight decay (max $1\times10^{-4}$). We follow the same policy as \cite{yolov8_ultralytics} loading weights of a detector pre-trained on the COCO dataset \cite{lin2015microsoft} using an exponential moving average on the weights for a more stable training evolution. %
Maximum VRAM and inference time were measured on an NVIDIA 1080Ti GPU.

\section{Experimental Results}\label{sec:results}

We validate  AuXFT on four benchmarks against some baseline architectures and compute-heavy upper bounds (we ignore FM-based strategies since their cost makes them unviable on-device). %
We choose two metrics for the evaluation, both averaged over 100 episodes, as common practice in FSL strategies: \map{ }(mean Average Precision 50-95) and retrieval accuracy.
\map{ } is the average of the IoU-thresholded mAPs from mAP50 to mAP95 (in steps of 5), each measuring the ratio of correctly predicted boxes with at least N\% IoU \cite{lin2015microsoft,yolov8_ultralytics}. The retrieval accuracy instead measures the fraction of boxes  correctly classified, regardless of localization.
We remark that %
both are computed on the fine-grained class set, which on average, contains 5 times more classes than the coarse set in the considered datasets. 

The quantitative results for one- and five-shot learning are reported in Table~\ref{tab:results:fine}, together with the average throughput, maximum VRAM usage, and model size. Here we report five groups of strategies: i) two baselines (non-distilled YOLOv8n); ii) two competing approaches (similar to hint-based knowledge distillation); iii) four implementations of our approach (changing $D_C$); %
iv) and two oracle upper bounds (joint use the object detector and the SSL pretrained extractor). 
Additionally, we report a global upper limit, which is the performance of the detector on the coarse class-set, and the relative performance of our method when compared to the \textit{Yolo Only (last feat.)} baseline and the \textit{Oracle (single fwd.)} upper bound.
The difference between the two baselines lies in how the vectors input to the FSL are generated: in \textit{last feat.}\ we use YOLO's last feature map only, while in \textit{concat.} we use all three feature maps and concatenate the results.
In the \textit{single fwd.} oracle setup we use the FSL in the same way as our approach, but substitute $\mathbf{R}_1$ with $\mathbf{O}$. In the other scenario, \textit{multiple fwd.}, we use the detector predictions to cut the original RGB image, rather than the feature maps, these sections are then fed into the oracle network to extract a vector embedding. Note that this strategy %
needs as many forward passes %
as the number of detected boxes. %
 
\begin{table}[t]
    \centering
    \caption{Results when pretraining class set matches the deployment set. Average \map{ }over 100 episodes ($\uparrow$).}
    \label{tab:results:pretrain}
    \begin{tabular}{c|c|cc}
        \multirow{2}{*}{Setup} & PerSeg (13/39) & \multicolumn{2}{c}{POD (15/45)} \\
        & 1-shot & 1-shot & 5-shot \\
        \midrule
        Yolo Only (last feat.) & 34.8\tiny{$\pm$3.0} & 22.1\tiny{$\pm$2.2} & 32.0 \\
        Yolo Only (concat.) & 38.2\tiny{$\pm$2.9} & 25.6\tiny{$\pm$2.4} & 30.3 \\
        \hdashline
        AuXFT (3x3 lin.) & 49.5\tiny{$\pm$3.2} & 33.0\tiny{$\pm$2.7} & 40.6 \\
        \hdashline
        Oracle (single fwd.) & 60.4\tiny{$\pm$3.2} & 42.3\tiny{$\pm$2.8} & 48.2 \\
        Oracle (multiple fwd.) & 61.2\tiny{$\pm$3.2} & 47.6\tiny{$\pm$2.9} & 61.9 \\
        \midrule[1pt]
        AuXFT Rel.\ Baseline & 142.2\% & 149.3\% & 126.9\% \\
        AuXFT Rel.\ Oracle & 82.0\% & 78.0\% & 84.2\%
    \end{tabular}
    \vspace{-1.5em}
\end{table}

Table \ref{tab:results:fine} shows how our strategy significantly improves performance over the baseline strategies and competitors using the unmodified feature space, regardless of the actual channel differential block implementation. Compared to the baseline, AuXFT gains an average of 10.5 mAP in PerSeg, 9.7 mAP in POD 1-shot scenario, and 9.4 in POD 5-shot.
This performance is a very close match with the oracle upper bound, where we achieve a consistent 80\% relative performance with less than a $1/3$ of the inference time and VRAM usage. Conversely, due to the scenario mismatch involved in iCubWorld and CORe50, the gains in retrieval accuracy are less impressive in absolute terms, but much more significant when compared to the upper bound. In these datasets, we achieve an astonishing $87.7\%$ of the oracle performance. %
Another interesting point regards the efficiency of our approach: with just a small increase of 0.3 ms/im in throughput and 65MB of VRAM %
compared to the baseline, %
AuXFT \textit{1x1 lin.}\ as well as \textit{3x3 lin.}\ increase the mAP by about 10 points on PerSeg and POD. Given that VRAM and inference time of the two approaches are nearly identical, we choose the \textit{3x3 lin.} as our reference implementation.

We remark that the performance in POD is lower than PerSeg due to the inherently more difficult setup. Indeed, POD has more classes and contains dark scenes.
As reported in Table~\ref{tab:results:fine} the global upper limit is 70.0 mAP, but the number rises to 79.0 when only bright scenes are considered (and it decreases to 61.5 for dark scenes only). This difference is reflected in the personalization as well, where our approach can achieve 33.9 (46.9) in bright scenes, on 1- (5-) shot, respectively. In dark scenes the performance is lower, falling to 31.0 (35.0). %

To further explore our approach we report some additional experiments in Tables \ref{tab:results:pretrain} and \ref{tab:results:coarse}. In the former we study the results when dataset-specific pretraining is employed, that is, the OpenImages subset is selected to match exactly the deployment class set.
In the latter, we explore the performance of our approach when coarse-level fallback is enabled in the FSL.

Table \ref{tab:results:pretrain} %
shows that using dataset-specific checkpoints for our approach gains on average 1.3 mAP compared to the unified pretraining results (Table~\ref{tab:results:fine}). %
This confirms the viability of pretraining the detector once for all considered scenarios. Even more, in the baseline and oracle runs the improvement is inconsistent: in the PerSeg dataset using a restricted class set leads to an average performance decrease of 2.4 mAP, conversely, in POD, we see an improvement of 1.4. This is probably due to the different number of personal instances per coarse class in the two datasets.

\begin{table}[t]
    \centering
    \caption{Results when both fine and coarse classes are predicted in FSL. Average \map{ }over 100 episodes ($\uparrow$).}
    \label{tab:results:coarse}
    \begin{tabular}{c|c|cc}
        \multirow{2}{*}{Setup} & PerSeg (13/39) & \multicolumn{2}{c}{POD (15/45)} \\
        & 1-shot & 1-shot & 5-shot \\
        \midrule
        Yolo Only (last feat.) & 19.9\tiny{$\pm$2.7} & 12.5\tiny{$\pm$1.9} & 19.3 \\
        Yolo Only (concat.) &  22.4\tiny{$\pm$2.8} & 14.1\tiny{$\pm$2.3} & 23.3 \\
        \hdashline
        AuXFT (3x3 lin.) & 34.5\tiny{$\pm$3.2} & 22.9\tiny{$\pm$2.7} & 34.1 \\
        \hdashline
        Oracle (single fwd.) & 46.7\tiny{$\pm$3.0} & 26.9\tiny{$\pm$2.6} & 38.5 \\
        Oracle (multiple fwd.) & 49.9\tiny{$\pm$2.9} & 33.6\tiny{$\pm$2.9} & 50.1 \\
        \midrule[1pt]
        AuXFT Rel. Baseline & 173.4\% & 183.2\% & 176.7\% \\
        AuXFT Rel. Oracle & 73.8\% & 85.1\% & 88.6\%
    \end{tabular}
    \vspace{-1em}
\end{table}

The results in Table~\ref{tab:results:coarse} confirm the effectiveness of our strategy, even in the challenging scenario where the FSL predictions belong to $\mathcal{C}$, which contains 51 and 60 classes, for PerSeg and POD, respectively.
In general, the experiments are consistent with all previous cases showing how our approach bridges the gap between the baseline and oracle architectures. %
Interestingly, in this scenario, the relative gain compared to the baseline is much more significant, with an average of 177.8\%. Comparing with the oracle we see a different behavior: in POD the performance increases (compared to results in Table~\ref{tab:results:fine}) by an average of 7.4\%, while in PerSeg we see a loss of 5\%.
We suppose this depends on the unbalanced nature of PerSeg's class set, which disproportionally penalizes our method when the coarse classes are enabled in the FSL. We remind the reader that PerSeg has multiple classes with only one instance (see Sec.~\ref{subsub:datasets}).

\begin{figure*}[t]
    \newcommand{\imgWidth}{.15\textwidth}
    \begin{subfigure}{\textwidth}
        \centering
        \begin{subfigure}{\imgWidth}
            \includegraphics[width=\textwidth]{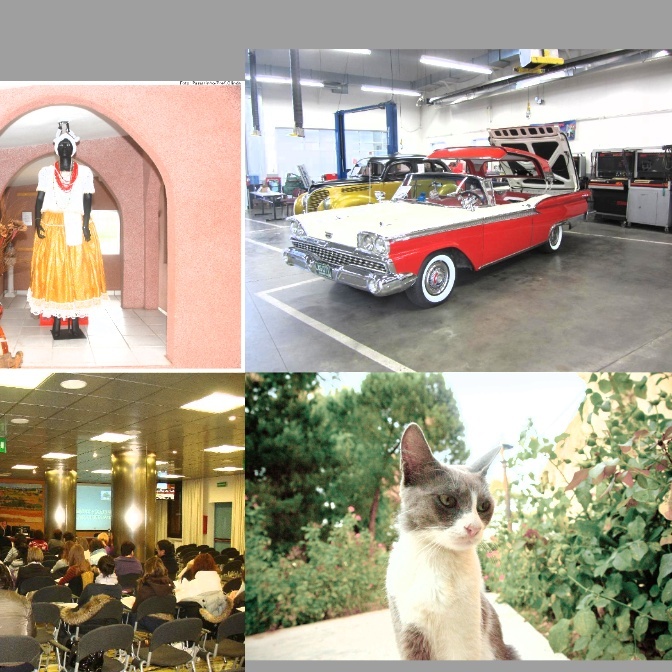}
            \caption{RGB Input}
        \end{subfigure}
        \begin{subfigure}{\imgWidth}
            \includegraphics[width=\textwidth]{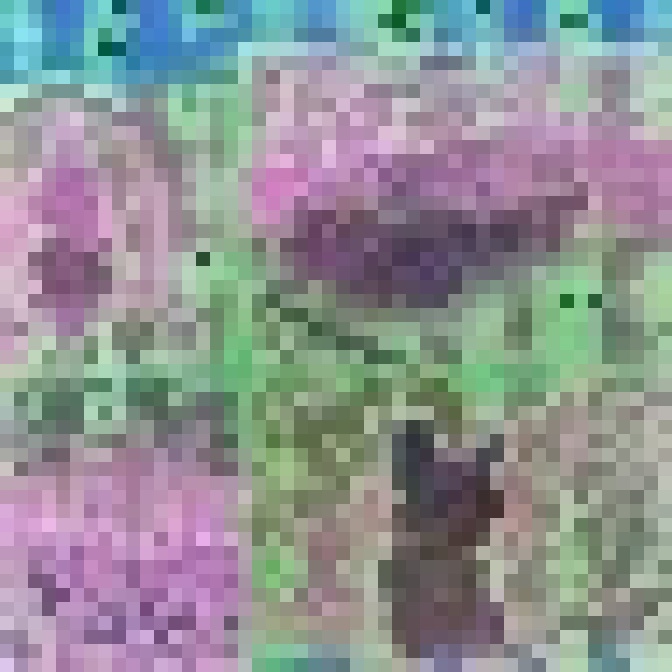}
            \caption{Oracle ($\mathbf{O}$)}
        \end{subfigure}
        \begin{subfigure}{\imgWidth}
            \includegraphics[width=\textwidth]{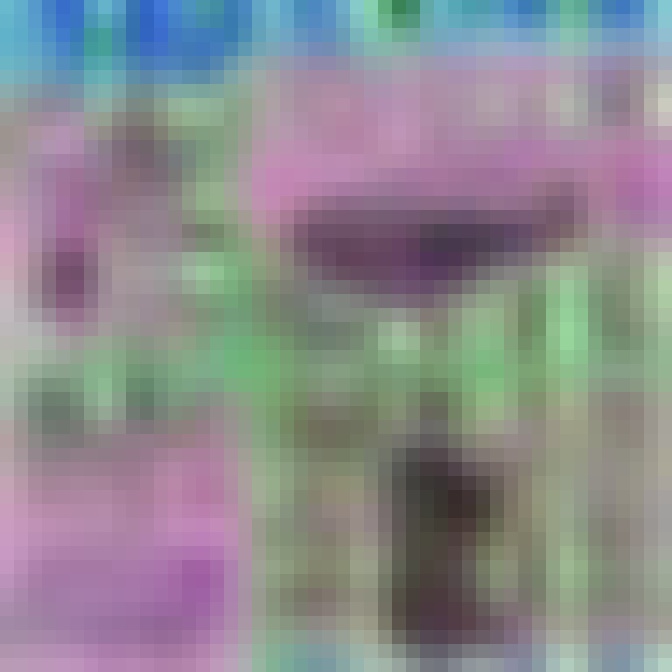}
            \caption{Partial I ($\mathbf{R}_3$)}
        \end{subfigure}
        \begin{subfigure}{\imgWidth}
            \includegraphics[width=\textwidth]{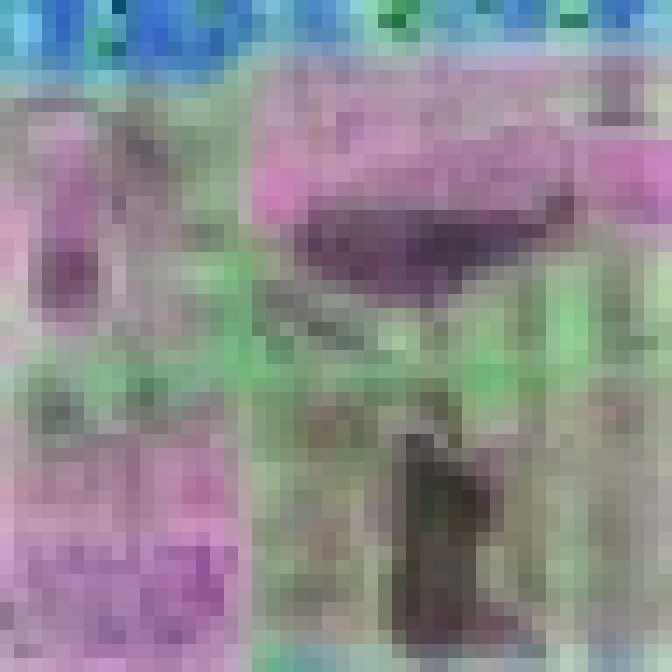}
            \caption{Partial II ($\mathbf{R}_2$)}
        \end{subfigure}
        \begin{subfigure}{\imgWidth}
            \includegraphics[width=\textwidth]{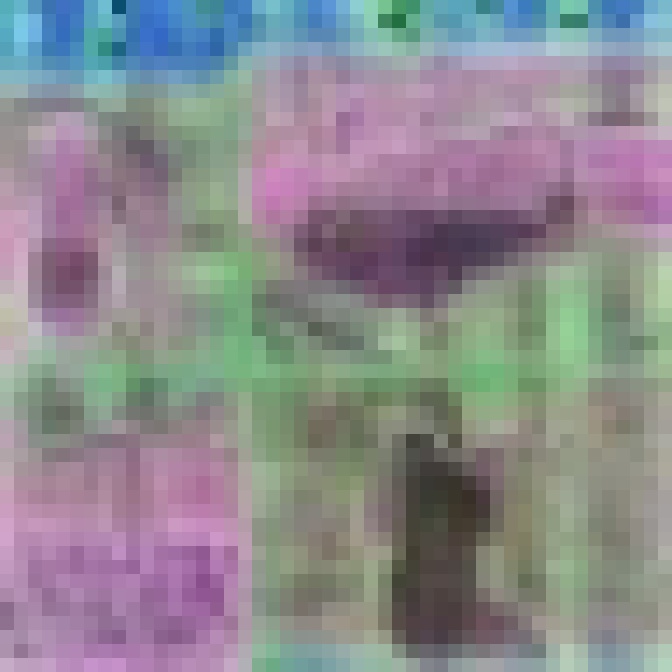}
            \caption{Full ($\mathbf{R}_1$)}
        \end{subfigure}
        \begin{subfigure}{\imgWidth}
            \includegraphics[width=\textwidth]{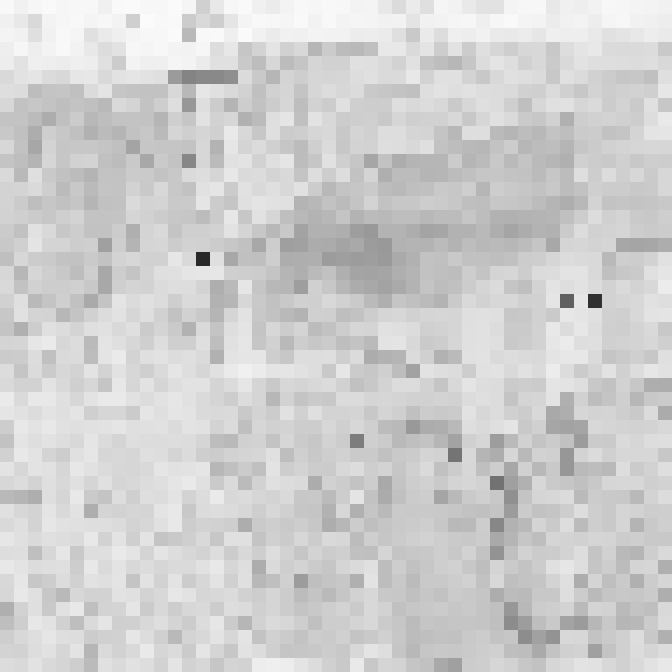}
            \caption{Cosine Sim.}
        \end{subfigure}
    \end{subfigure}
    \caption{Auxiliary Features visualization. To generate the colors, during training, we fit an embedding of the oracle features $\mathbf{O}$ in $[0,1]^3$ (RGB space), the same mapping is used to color $\mathbf{R}_i$. The cosine similarity is computed pointwise.}
    \label{fig:features}
    \vspace{-1em}
\end{figure*}

\begin{figure}[t]
    \centering
    \includegraphics[width=\linewidth]{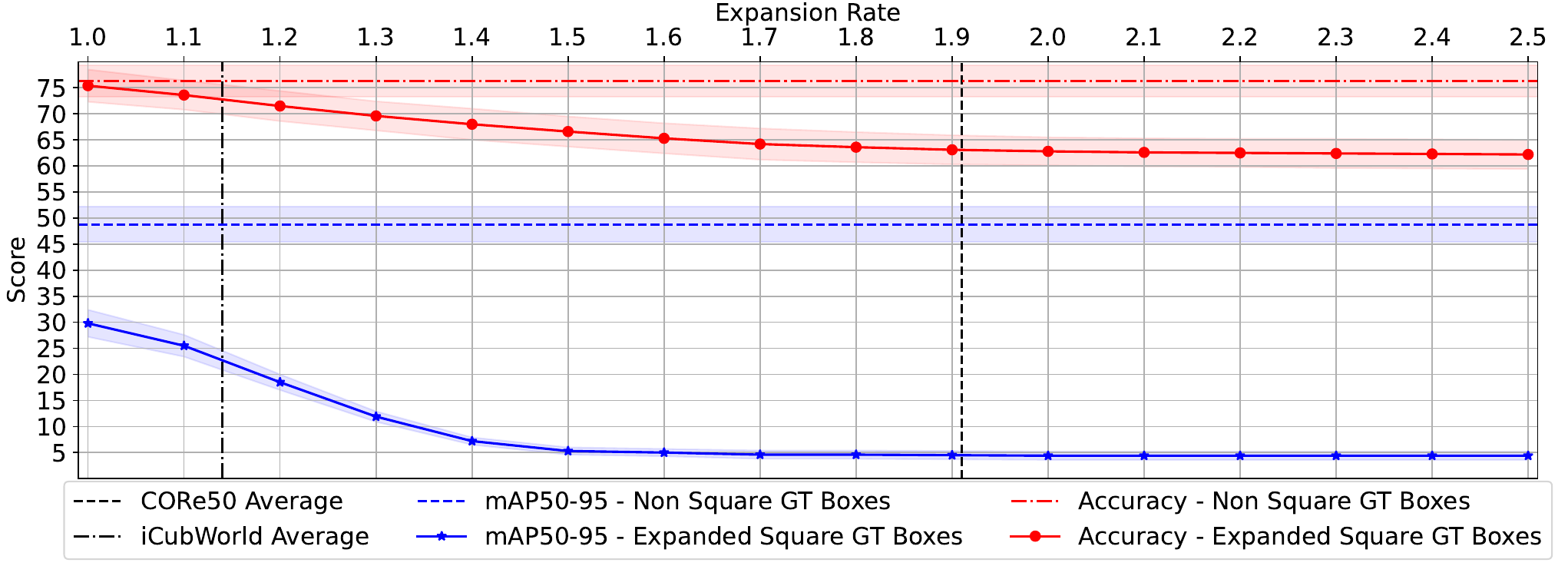}
    \caption{Personalization performance varying the ground truth boxes padding size on PerSeg.}
    \label{fig:ablation:boxes}
    \vspace{-1em}
\end{figure}

\section{Ablation}\label{sec:ablation}

In this section we analyze in more detail some of the design choices made for the proposed architecture: i) in Table \ref{tab:ablation:proto} we focus on the distance functions involved in the FSL, showing how their combined use leads to a more stable performance; ii) in Table \ref{tab:ablation:fsl} we show the effect of changing the FSL, comparing our conditional majority voting strategy to other protonet-based strategies; %
and iii) in Table~\ref{tab:ablation:features} we show the performance of the FSL when different residual maps $\mathbf{R}_i$ are used (as opposed to $\mathbf{R}_1$). 
For this study, in Fig.~\ref{fig:features} we also provide some qualitative results visually showing the improvement in reconstruction. 
We show a colorized version of the oracle ($\mathbf{O}$) and auxiliary ($\mathbf{R}_i$) feature spaces, together with the original input image (augmented according to the training policy) and the pointwise cosine similarity between $\mathbf{R}_1$ and $\mathbf{O}$.
The color is generated according to a non-linear embedding which is fit on the oracle features at training time.
The channel-dense features into $[0,1]^3$ (RGB space) are mapped using a 1x1 convolution with sigmoid activation function, the embedding is then expanded to the original space with another 1x1 convolution and a reconstruction loss between the original and reconstructed features is minimized.

\begin{table}[t]
    \centering
    \caption{Effect of different distances in the FSL accuracy. \textbf{Best} in bold, \underline{second best} underlined.}
    \label{tab:ablation:proto}
    \begin{tabular}{ccc|cc}
        \multirow{2}{*}{cos} & \multirow{2}{*}{$\ell_1$} & \multirow{2}{*}{$\ell_2$} & \multicolumn{2}{c}{mAP ($\uparrow$)} \\
         & & & Coarse \xmark & Coarse \cmark \\
        \hline
        \cmark & \xmark & \xmark & 48.3\tiny{$\pm$3.3} & \textbf{34.6\tiny{$\pm$3.2}} \\
        \xmark & \cmark & \xmark & 48.6\tiny{$\pm$3.2} & 33.8\tiny{$\pm$3.1} \\
        \xmark & \xmark & \cmark & \textbf{48.9\tiny{$\pm$3.3}} & 34.2\tiny{$\pm$3.2} \\
        \hdashline
        \cmark & \cmark & \xmark & 48.4\tiny{$\pm$3.3} & \textbf{34.6\tiny{$\pm$3.2}} \\
        \cmark & \cmark & \cmark & \underline{48.8\tiny{$\pm$3.4}} & \underline{34.5\tiny{$\pm$3.2}} \\
    \end{tabular}
    \vspace{-1em}
\end{table}

For the analysis in Table \ref{tab:ablation:proto} we consider three distance functions: cosine, $\ell_1$ and $\ell_2$. %
The table shows how different distance functions are better depending on whether the coarse classes are considered or not (cosine is best when they are enabled, $\ell_2$ otherwise). For this reason, we employ all three distances together, achieving a good compromise (second-best performance in both cases).

\begin{table}[t]
\begin{minipage}{.55\linewidth}
    \centering
    \caption{Comparisons between different Few Shot Learners.}
    \label{tab:ablation:fsl}
    \resizebox{!}{.2\linewidth}{%
    \setlength{\tabcolsep}{2pt}
    \begin{tabular}{c|cc}
        \multirow{2}{*}{Model} & \multirow{2}{*}{mAP ($\uparrow$)} & Time \\
        & & [ms/im] ($\downarrow$) \\
        \hline
        Protonet-cos \cite{snell2017prototypical} & 50.8\tiny{$\pm$3.2} & 144.8\tiny{$\pm$6.7} \\
        Protonet-$\ell_2$ \cite{snell2017prototypical} & \textbf{51.0\tiny{$\pm$3.5}} & 96.1\tiny{$\pm$5.9} \\
        SimpleShot \cite{wang2019simpleshot} & 50.8\tiny{$\pm$3.2} & 94.1\tiny{$\pm$5.8} \\
        \hline
        Ours & 48.8\tiny{$\pm$3.4} & \textbf{32.6\tiny{$\pm$5.1}} \\
    \end{tabular}}
\end{minipage}\hspace{1em}
\begin{minipage}{.4\linewidth}
    \centering
    \caption{Results on residual chain shortcuts.}
    \label{tab:ablation:features}
    \begin{tabular}{c|c}
        Features & mAP ($\uparrow$) \\
        \hline
        Partial I ($\mathbf{R}_3$) & 48.4\tiny{$\pm$3.5} \\
        Partial II ($\mathbf{R}_2$) & 48.7\tiny{$\pm$3.4}\\
        Full ($\mathbf{R}_1$) & \textbf{48.8\tiny{$\pm$3.4}}
    \end{tabular}
\end{minipage}
\vspace{-1em}
\end{table}

We compare the proposed FSL strategy  with state-of-the-art competitors in Table \ref{tab:ablation:fsl}, measuring both personalization accuracy and time complexity.
The table shows how our conditional approach leads to a slight decrease in performance (2.2 mAP compared to the best strategy), but improves the computation time about $3\times$. Such a trade-off is fundamental for robotic applications where many vision tasks require real-time performance.

Fig.~\ref{fig:ablation:boxes} reports a study on the padding of GT boxes, which shows how sensitive the mAP is to its presence. %
The results justify the choice of retrieval accuracy made for iCubWorld and CORe50, given that the metric is much more resilient to changes in box size. For further confirmation, we measured the mAP of iCubWorld and obtained 20.9.
This is very close to the expected value from the figure (i.e., around 22.5).

The final ablation study we report refers to the residual chain of the auxiliary features: we  investigated the performance attained by our method when the partial reconstructions $\mathbf{R}_i$ were used instead of $\mathbf{R}_1$. The results shown in Table~\ref{tab:ablation:features} confirm the improvements brought by the chain: using only $\mathbf{R}_n$ ($n=3$ in our setup) results in a total of 48.4 mAP which increases to 48.7 when two maps are merged. The best performance is improved again when $\mathbf{R}_1$ is also used, slightly raising to 48.8.

\section{Conclusions}\label{sec:conclusions}

In this work, we presented \textit{AuXFT} a Few-Shot Instance-Level Personalized Object Detection architecture, which translates the features of a CNN-based detector into an auxiliary feature space where cross-architecture knowledge translation is possible. We have designed a framework for personalizing coarse-level detection to instance-level classes using a prototypical-based FSL that receives input vectors generated by our detection-driven feature pooling. We have validated the performance of our architecture on four different personalized object detection datasets, achieving significant gains in all of them. 
In the future, we plan to investigate the approach using different detectors and oracle extractors to verify the generalizability of the strategy and its suitability for deployment on resource-constrained robotic devices. We also plan to test new mapping functions in the channel differential block and whether some parametric resampling strategies (such as transposed convolutions) can improve performance when added to the spatial differential module.

\bibliographystyle{ieeetran}
\bibliography{strings_short,refs}
\end{document}